\documentclass[conference]{IEEEtran}
\IEEEoverridecommandlockouts

\usepackage{cite}
\usepackage{amsmath,amssymb,amsfonts}
\usepackage{algorithmic}
\usepackage{algorithm}
\usepackage{graphicx}
\usepackage{textcomp}
\usepackage{xcolor}
\usepackage{booktabs}
\usepackage{array}
\usepackage{url}
\usepackage{hyperref}
\usepackage{multirow}
\usepackage{balance}
\usepackage{enumitem}
\usepackage{microtype}

\hypersetup{colorlinks=true,linkcolor=blue,citecolor=blue,urlcolor=blue}

\newcommand{\etal}{\textit{et al.}}

\newcommand{\shield}{\textsc{LiDAR-SHIELD}}

\begin{document}

\title{LiDAR-SHIELD: A Fully Dockerized Adversarial Robustness Assessment
Framework for LiDAR-Based 3D Perception Models
\thanks{This work has received funding from the EU’s Horizon Europe research and innovation programme in the frame of the CoEvolution project “A comprehensive trustworthy framework for connected machine learning and secured interconnected AI solutions” under the Grant Agreement No 101168560.
}}

\author{
\IEEEauthorblockN{
Alexandros Gkillas$^{2}$, Ioulia Kapsali$^{1}$, Christos Anagnostopoulos$^{1}$, and Aris S. Lalos$^{1}$
}
\IEEEauthorblockA{$^1$Industrial Systems Institute, Athena Research Center, Patras Science Park, Greece\\
$^2$AviSense.AI, Patras Science Park, Greece\\
Emails: \{anagnostopoulos, gkillas\}@avisense.ai, \{ikapsali,lalos\}@athenarc\}.gr 
}}

\maketitle

\begin{abstract}
LiDAR-based 3D perception underpins autonomous driving and robotics, yet deploying these models safely requires stress-testing them against adversarial inputs before real-world use. Although state-of-the-art adversarial attacks exist, their implementations are often fragmented across individual repositories, each tied to a specific architecture or task and requiring manual environment reconstruction and model-specific integration code. We present \shield{}, a unified and fully Dockerized framework that reverses this workflow: rather than installing the target model inside the toolkit, \shield{} exposes any model through three lightweight HTTP endpoints, namely \texttt{/info}, \texttt{/forward}, and \texttt{/gradient}. As a result, connecting a new model or dataset requires only a single loader function or adapter. The framework supports both 3D semantic segmentation and 3D object detection through the same engine-service protocol, integrating a task-aware suite of state-of-the-art attacks, including PGD and Square Attack for segmentation, and point perturbation, point removal, point injection, and LiDAttack for detection. This enables systematic exposure of corner cases and vulnerabilities across heterogeneous LiDAR backbones. The LiDAR-Shield codebase is publicly available at
\url{https://github.com/alexandrosgk/LiDAR-shield}. .

\end{abstract}

\begin{IEEEkeywords}
LiDAR perception, adversarial robustness, 3D semantic segmentation, 3D
object detection, adversarial attacks, autonomous driving.
\end{IEEEkeywords}

\section{Introduction}
\label{sec:intro}

LiDAR is a primary sensing modality in autonomous driving and robotics, providing accurate metric geometry, long-range structure, and stable
performance across illumination conditions that often challenge camera-only
systems~\cite{Geiger2012CVPR, Waymo2021Report}.
Two perception tasks dominate this setting.
\textbf{3D object detection} localizes and classifies dynamic agents such as
pedestrians, cyclists, and vehicles, providing the instance-level signals
needed for prediction and collision avoidance~\cite{
Yin2021CenterBased}.
\textbf{3D semantic segmentation} assigns dense semantic labels to the sensed
scene, enabling drivable-area reasoning, obstacle delineation, and map-aware
scene understanding~\cite{Behley2019ICCV, Milioto2019IROS, Cheng2022TIP,Gkillas2026Conditional}.
However, the same
differentiable pipelines that enable strong predictive
performance also expose a broad adversarial surface,
and before
deployment these models must be stress-tested against adversarial inputs
to identify corner cases and vulnerabilities that clean
benchmarks do not expose.

Adversarial perturbations in LiDAR can suppress detections,
distort box localization, or erase safety-critical semantics
under imperceptible or physically plausible modifications to
the input. LiDAR attacks span \textbf{white-box} methods that
optimize perturbations using model
gradients~\cite{Madry2018ICLR}, \textbf{black-box} methods that rely only on
model outputs~\cite{Brendel2018ICLR, Chen2020RayS}, and
\textbf{physical} attacks that alter the sensed
scene;
perturbations may act in projected range-image
space~\cite{Zhu2021CVPR, Liu2022Robustness, Gkillas2025Efficient}, directly on raw
point clouds~\cite{Wicker2019ICLR, Xiang2019AdvPC}, or at
the sensor level.
Because detection and segmentation share the same upstream
LiDAR measurement, input-space perturbations can compromise
both tasks simultaneously.


Yet robustness evaluation infrastructure has not kept pace with the attack
literature.
General adversarial toolkits such as ART~\cite{Nicolae2018ART},
Foolbox~\cite{Rauber2017Foolbox}, CleverHans~\cite{Papernot2018CleverHans}, and
secml~\cite{Pintor2022secml} follow a \emph{library-centric}
integration model: evaluating a model requires installing it
within the toolkit's codebase, adapting its interface to the
library's abstractions, and resolving the resulting dependency
conflicts. 
Beyond this integration barrier, none provides native support
for LiDAR point cloud inputs or 3D perception outputs. 
At the LiDAR perception level, existing work either addresses
a single task~\cite{Zhang2024Comprehensive}, targets natural
corruptions rather than adversarial
attacks~\cite{Kong2023Robo3D}, or releases attack code
entangled with a specific architecture and software
stack, requiring manual environment reconstruction and
model-specific glue code to reuse.
As a result, a researcher wishing to stress-test a LiDAR backbone 
across both perception tasks
still lacks a single systematic and reproducible tool.

We address this gap with \shield{}, a unified adversarial robustness
assessment framework for LiDAR perception.
Unlike library-centric toolkits,
\shield{} adopts a \emph{service-oriented} design: the model remains in
its native environment and communicates with the assessment
engine by simply sending data and receiving outputs or
gradients through three lightweight HTTP endpoints.
Connecting a new model or dataset requires only a single
loader function or adapter, without any model code entering
the framework or dependency conflicts arising.
The architecture pushes two practical sources of heterogeneity
to thin boundary components: a data adapter handles dataset
variability (file layout, calibration, annotation structure),
while a model service handles model variability (dependencies,
preprocessing, loss definitions).
The resulting system is fully Dockerized and supports both 3D semantic
segmentation and 3D object detection within the same architectural
abstraction.
Our contributions are:
\begin{enumerate}[leftmargin=*, noitemsep, label=\textbf{(\arabic*)}]
  \item We introduce a \textbf{model-agnostic assessment architecture} that
    decouples adversarial evaluation from the target model through a minimal
    REST interface, eliminating manual environment reconstruction and
    model-specific glue code.
  \item We present a \textbf{unified workflow for both 3D semantic
    segmentation and 3D object detection}, enabling direct robustness
    evaluation across the two dominant LiDAR perception tasks within a single
    framework.
  \item We implement a \textbf{task-aware suite of state-of-the-art attacks}
    capable of exposing corner cases and vulnerabilities across heterogeneous
    LiDAR backbones: PGD and Square Attack for segmentation,
    and point perturbation, point removal, point injection, and
    LiDAttack for detection.
  \item We provide a \textbf{fully Dockerized deployment model} that isolates
    data handling, model execution, and attack orchestration, reducing
    the engineering overhead of adversarial robustness testing to a
    plug-and-play systems component.
\end{enumerate}

\section{Related Work}
\label{sec:related}

\textbf{Adversarial attacks on LiDAR perception.}
Prior work has established that LiDAR perception models are vulnerable under
multiple threat models.
White-box digital perturbations can suppress detections or
induce localization errors in 3D
detectors~\cite{ Xie2020Adversarial}, and
adversarial examples transfer across LiDAR
architectures~\cite{Zhu2021CVPR}.
Recent work has further examined the vulnerability of LiDAR segmentation
pipelines to digital perturbations~\cite{Liu2022Robustness, Gkillas2025Efficient}.
At the point-cloud level, point deletions suffice to fool 3D
models~\cite{Wicker2019ICLR}, and transferable perturbations
on raw 3D clouds have also been
demonstrated~\cite{Xiang2019AdvPC}.
More recently, Zhang~\etal~\cite{Zhang2024Comprehensive} provided a
comprehensive study of adversarial attacks on LiDAR-based 3D detectors, and
Chen~\etal~\cite{Chen2025LiDAttack} proposed LiDAttack, a robust black-box
attack targeting LiDAR-based object detection.
However, these attack implementations do not provide a reusable evaluation infrastructure that
spans both detection and segmentation.

\textbf{LiDAR perception models.}
LiDAR perception has diversified into heterogeneous model families for both
segmentation and detection.
For segmentation, RangeNet++~\cite{Milioto2019IROS}, CENet~\cite{Cheng2022TIP},
FIDNet~\cite{Zhao2021IROS}, and LENet~\cite{Feng2022LENet} represent distinct
design points in efficient dense scene understanding.
For detection, PointPillars~\cite{Lang2019CVPR}, VoxelNet~\cite{Zhou2018CVPR},
SECOND~\cite{Yan2018SENSORS}, and CenterPoint~\cite{Yin2021CenterBased}
illustrate the diversity of pillar-based, voxel-based, and center-based 3D
detection pipelines.
This heterogeneity motivates a service abstraction that can
accommodate arbitrary backbones without modifying the
assessment core.

\textbf{Robustness toolkits and benchmarks.}
General adversarial toolkits such as ART~\cite{Nicolae2018ART},
Foolbox~\cite{Rauber2017Foolbox}, and CleverHans~\cite{Papernot2018CleverHans} and secml~\cite{Pintor2022secml} require the target model to
be embedded within their own codebase, coupling the evaluation
pipeline to a specific framework and dependency stack.
In the 3D domain, Ji~\etal~\cite{Ji2023Benchmarking} benchmark
adversarial robustness for point cloud recognition, but target object classification rather than the detection
and segmentation tasks that underpin autonomous driving.
At the LiDAR perception level, Zhang~\etal~\cite{Zhang2024Comprehensive} benchmark adversarial robustness across multiple 3D detectors, but their
study is confined to detection and does not extend to segmentation.
Robo3D~\cite{Kong2023Robo3D} covers both 3D detectors and segmentors under a
unified evaluation protocol, but focuses on natural corruptions 
rather than adversarial attacks.
Beyond task and domain coverage, LiDAR attack implementations
released alongside individual papers are each tied to a
specific codebase, dependency stack, and model interface,
making reuse costly: it requires reconstructing the original
environment and writing model-specific glue code for every
new target. Taken together, no existing tool provides unified,
Dockerized adversarial evaluation spanning both LiDAR
segmentation and detection, the gap that \shield{} is
designed to fill.

\section{Method}
\label{sec:solution}

\begin{figure*}
  \centering
  \includegraphics[width=0.5\linewidth]{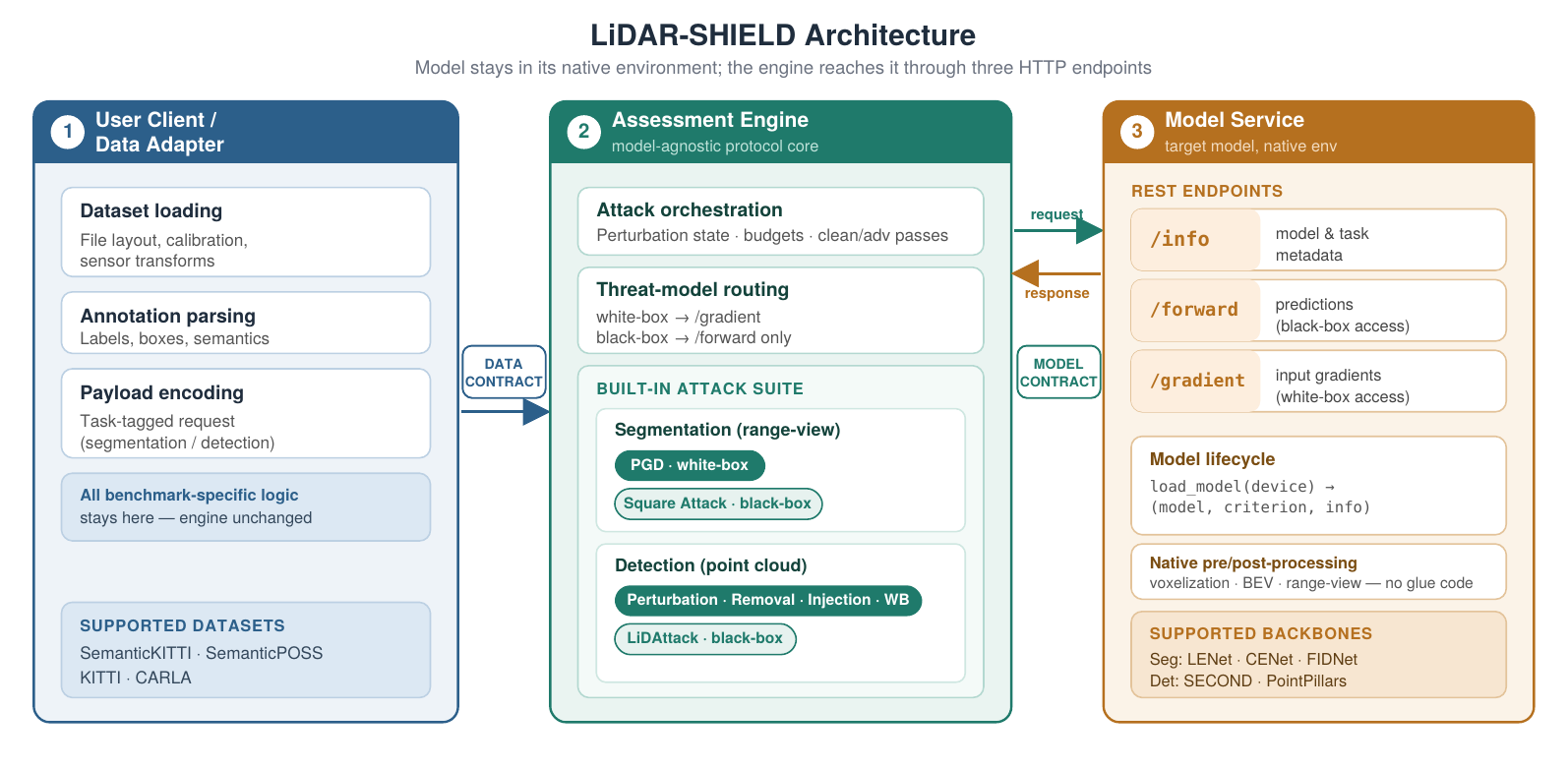}
 \caption{\shield{} architecture. The target model stays in its native
  environment and is reached through three HTTP endpoints (\texttt{/info},
  \texttt{/forward}, \texttt{/gradient}). A \emph{data contract} normalizes
  dataset samples on the left and a \emph{model contract} normalizes backbone
  access on the right, leaving the engine to run a task-aware, model-agnostic
  attack pipeline. White-box attacks use \texttt{/gradient}; black-box attacks
  use \texttt{/forward} only.}
     \label{fig:arch}
\end{figure*}

The central design question behind \shield{} is how to
evaluate adversarial robustness without inheriting the full
software stack of each target model. The answer is to isolate
the two dominant sources of variability, dataset
variability and model variability, at the boundary of the
system rather than absorbing them into the attack pipeline.
\shield{} factors the robustness evaluation problem into three
Dockerized components: a \emph{User Client / Data Adapter} that
owns dataset interaction, an \emph{Assessment Engine} that owns
the adversarial protocol, and a \emph{Model Service} that owns
network execution. This decomposition isolates dataset-specific
assumptions from model-specific assumptions and leaves the
engine with a single role: executing a task-aware but
model-agnostic attack pipeline. The framework is defined by two
contracts: a data contract that maps dataset samples into
engine-consumable payloads, and a model contract that maps
arbitrary LiDAR backbones into a common service interface,
allowing new datasets and models to plug into the same engine
without modifying the assessment core.
Fig.~\ref{fig:arch} illustrates the resulting architecture.

\subsection{Architectural Overview}

At a high level, \shield{} is organized as a left-to-right execution pipeline.
The \textbf{User Client / Data Adapter} converts dataset-native LiDAR samples
into task-tagged requests.
The \textbf{Assessment Engine} receives these requests, instantiates the
selected attack, and manages the interaction loop with the target model.
The \textbf{Model Service} wraps the network and exposes only the
operations required by the engine.
The assessment engine is therefore the protocol layer between two normalized
boundaries: on the left, dataset-specific samples are mapped into a uniform
payload format; on the right, backbone-specific execution is mapped into a
fixed service API.
\shield{} does not attempt to standardize how datasets are
stored or how models are implemented; it standardizes only how data enters the
engine and how the engine queries the model.

\subsection{Components}


\textbf{User Client / Data Adapter.}
The client/data-adapter is the ingestion boundary of the framework,
responsible for loading LiDAR samples and their task annotations,
arranging them into the model's expected input structure, and packaging
them into a transport-safe request.
All benchmark-specific logic remains here, including file layout, sequence
handling, calibration usage, and annotation parsing, so that
connecting a new dataset does not require changing the engine.
The connection mechanism is intentionally thin: to support a new dataset,
one implements only this adapter layer so that dataset-native samples are
emitted in the engine's data contract.
\textbf{Assessment Engine.}
The assessment engine is the methodological core of \shield{}, implementing
the adversarial protocol itself: selecting the attack,
maintaining perturbation state, coordinating clean and adversarial passes,
and executing the task-aware control logic associated with the chosen
threat model.
This ensures that attack behavior is implemented once and reused across
heterogeneous LiDAR models, regardless of their underlying codebase.
\textbf{Model Service.}
The model service is the execution boundary around the target perception
network, encapsulating model construction, checkpoint loading, task-loss
definition, and model-specific preprocessing or postprocessing,
so that the assessment engine never imports
repository-local 
code.
To connect a new backbone, the user implements a lightweight loader entry point, \texttt{load\_model(device)}, 
returning \texttt{(model, criterion, info, device)},
where \texttt{criterion} is the task loss and \texttt{info} is a metadata
dictionary containing model identity, number of classes, input shape, and
other task-level properties needed by the engine.

\subsection{Interface Contract}

The architecture is governed by two narrow contracts.
The \textbf{data contract} defines what the engine expects
from the client/data-adapter: a task-tagged sample payload
containing the encoded model input together with the
supervision and metadata needed by the selected attack path. This contract is agnostic to the data source: once the
payload is formed, the engine consumes it directly without
parsing raw dataset artifacts. The \textbf{model contract} defines what the engine expects
via three endpoints: \texttt{/info} returns model and task
metadata; \texttt{/forward} returns task-native predictions;
and \texttt{/gradient} returns the input gradient required
by white-box attacks. This minimal contract suffices for both segmentation and
detection: once a backbone answers these three calls, the
engine can attack it without importing its native codebase.

\subsection{Engine Workflow}

Once a payload reaches the assessment engine, the remainder of the
procedure is uniform.
The engine first queries \texttt{/info} once per model session to
retrieve the metadata needed to instantiate the appropriate task-aware
processing path.
Given a sample $(\mathbf{x}, \mathbf{y})$, it next obtains a clean
reference prediction through the model service, instantiates the
requested attack under the chosen threat model, and drives the
interaction loop until an adversarial input
$\mathbf{x}_{\mathrm{adv}}$ has been produced. The attacked sample is then executed through the same service boundary and the resulting task-native outputs are collected for downstream evaluation. Because the workflow is defined entirely over the two
contracts, the same procedure applies regardless of which dataset
adapter or backbone is in use. The distinction between black-box and white-box settings is handled entirely inside this workflow: in black-box mode the engine relies only
on \texttt{/forward} outputs, while in white-box mode it additionally
invokes \texttt{/gradient}. Crucially, the orchestration layer is
unchanged in both cases, only the attack operator and the type of
model access it requests differ, which is what allows \shield{} to
support multiple threat models without duplicating the surrounding
systems logic.






\subsection{Supported Perception Tasks}
\shield{} supports both perception tasks through the same
service protocol and attack orchestration logic. For
segmentation, the service accepts the model's native LiDAR
input and returns dense predictions; for detection, it accepts
the detector's native input and returns structured object
hypotheses. The attack layer is task-aware, but the
engine-service interface is shared, this is what allows
\shield{} to remain unified at the systems level while
matching the geometry and failure modes of both tasks.

\section{Adversarial Attacks}
\label{sec:attacks}
\shield{} instantiates attacks at the task level rather than enforcing a
single perturbation primitive across all LiDAR models.
Throughout this section, a LiDAR
scan is represented as a point cloud $P=\{\mathbf{p}_i \in \mathbb{R}^{3}\}_{i=1}^{N}$.

\subsection{Segmentation Attack Suite}
\label{subsec:seg_attacks}

For LiDAR semantic segmentation, \shield{} applies adversarial perturbations
directly to the projected range-view representation consumed by the segmentation
backbone. Let
$\mathbf{Y}\in\mathbb{R}^{H\times W}$ denote the clean range-view input
tensor, where $H$ is the number of vertical LiDAR beams and $W$ is the
horizontal angular resolution. Let $\mathbf{Z}$ denote the corresponding
ground-truth semantic label map. The adversarial range-view input is obtained
by adding a  perturbation $\boldsymbol{\Delta}$ to the clean input:
\begin{equation}
\mathbf{Y}_{\mathrm{adv}}
=
\mathbf{Y}
+
\boldsymbol{\Delta},
\qquad
\|\boldsymbol{\Delta}\|_{p}
\leq
\epsilon,
\label{eq:seg_adv_input}
\end{equation}
where $\epsilon$ controls the perturbation budget. The attack objective is to
degrade segmentation performance by maximizing the segmentation loss under the
perturbation constraint:
\begin{equation}
\max_{\|\boldsymbol{\Delta}\|_{p}\leq\epsilon}
\mathcal{L}_{\mathrm{seg}}
\left(
f_{\theta}^{\mathrm{seg}}(\mathbf{Y}+\boldsymbol{\Delta}), \mathbf{Z}
\right),
\label{eq:seg_adv_objective}
\end{equation}
where $f_{\theta}^{\mathrm{seg}}$ is the target segmentation model and
$\mathcal{L}_{\mathrm{seg}}$ denotes the segmentation loss.

\paragraph{White-box attack (PGD)}
We instantiate the above objective using Projected Gradient Descent (PGD)~\cite{Madry2018ICLR}
via the \texttt{/gradient} endpoint. Starting from
$\mathbf{Y}^{0}=\mathbf{Y}$, each iteration is given by
\begin{equation}
\mathbf{Y}^{t+1}
=
\Pi_{\mathcal{B}_{p}(\mathbf{Y},\epsilon)}
\left(
\mathbf{Y}^{t}
+
\alpha\,
\mathcal{N}_{p}
\left(
\nabla_{\mathbf{Y}^{t}}
\mathcal{L}_{\mathrm{seg}}
\left(
f_{\theta}^{\mathrm{seg}}(\mathbf{Y}^{t}), \mathbf{Z}
\right)
\right)
\right),
\label{eq:pgd_seg}
\end{equation}
where $\alpha$ is the step size,
$\Pi_{\mathcal{B}_{p}(\mathbf{Y},\epsilon)}(\cdot)$ projects onto
the $\ell_p$ ball of radius $\epsilon$ 
and $\mathcal{N}_{p}(\cdot)$ is the norm-specific gradient
normalization for $p\in\{2,\infty\}$.

\paragraph{Black-box attack}
To assess robustness when gradient access is unavailable, \shield{}
additionally applies Square Attack~\cite{Andriushchenko2020SquareAttack}
as a black-box baseline. Square Attack operates
through \texttt{/forward} only, perturbing random
square-shaped regions of the input tensor via query-efficient
random search under an $\ell_\infty$ constraint, without
any gradient access.


\subsection{Detection Attack Suite}
\label{subsec:det_attacks}

For 3D object detection, \shield{} adopts perturbation operators that are
better aligned with sparse geometric evidence.
The ground-truth object annotations are denoted by $G$. The objective 
is to construct an adversarial point cloud
$P_{\mathrm{adv}}$ that increases the detection loss while preserving the
geometric plausibility of the original scene

\paragraph{White-box attacks}
\textbf{Point perturbation} displaces existing points by
small coordinate offsets $\boldsymbol{\delta}=
\{\boldsymbol{\delta}_i\}_{i=1}^{N}$:
\begin{equation}
P_{\mathrm{adv}} = \left\{\mathbf{p}_i+\boldsymbol{\delta}_i
\right\}_{i=1}^{N},
\label{eq:point_perturbation}
\end{equation}
under the constrained objective:
\begin{equation}
\max_{\{\boldsymbol{\delta}_i\}_{i=1}^{N}}
\mathcal{L}_{\mathrm{det}}
\!\left(f_{\theta}^{\mathrm{det}}(P_{\mathrm{adv}}),G\right)
\;\mathrm{s.t.}\;
\|\boldsymbol{\delta}_i\|_2 \leq \epsilon,\ \forall i,
\label{eq:point_perturbation_objective}
\end{equation}
where $f_{\theta}^{\mathrm{det}}$ is the target detector,
$\mathcal{L}_{\mathrm{det}}$ denotes the detection loss, and $\epsilon$
controls the maximum allowable displacement per point. This objective is
optimized iteratively via PGD~\cite{Madry2018ICLR}, with gradients obtained
through the \texttt{/gradient} endpoint.
\textbf{Point removal} deletes informative returns
($P_{\mathrm{adv}} \subset P$) in order of decreasing
saliency, where each point is scored by the $\ell_2$ norm
of its detection loss gradient
$\|\partial\mathcal{L}_{\mathrm{det}}(P,G)/\partial\mathbf{p}_i\|_2$,
excluding already-removed points from subsequent gradient
computations.
\textbf{Point injection} inserts adversarial point set $Z=\{\mathbf{z}_k\in\mathbb{R}^{3}\}_{k=1}^{K}$: 
\begin{equation}
P_{\mathrm{adv}} = P \cup Z,
\label{eq:point_injection}
\end{equation}
initialized at highest-saliency locations and optimized under
a Chamfer distance constraint; only $Z$ is updated during
optimization.
These operators follow established point-manipulation
literature~\cite{Wicker2019ICLR, Xiang2019AdvPC,
Xie2020Adversarial}.


\paragraph{Black-box attack}
To assess robustness when gradient access is unavailable, \shield{}
incorporates LiDAttack~\cite{Chen2025LiDAttack} as the black-box baseline
for 3D object detection. LiDAttack operates through
\texttt{/forward} only, using a genetic algorithm with
simulated annealing to find adversarial point-cloud
modifications without gradient access.


\subsection{Attack Comparison}
\label{subsec:attack_comparison}

Table~\ref{tab:attacks} summarises the attack portfolio supported by
\shield{}. Each task is covered by both a white-box and a black-box
operator: white-box
attacks establish an upper bound on adversarial risk under full model
access, while black-box attacks probe robustness under the more realistic
assumption that gradients are unavailable. Attack operators are chosen to match the input
space of each task, range-view tensors for segmentation
and raw point clouds for detection, allowing \shield{} to
support both tasks through the same engine-service interface
without sacrificing attack fidelity.

\begin{table}
\caption{Task-specific attack support in \shield{}.}
\label{tab:attacks}
\centering
\resizebox{\linewidth}{!}{%
\begin{tabular}{llll}
\toprule
\textbf{Task} & \textbf{Attack} & \textbf{Access} 
& \textbf{Perturbation Space} \\
\midrule
3D segmentation & PGD & White-box & Range-view tensor \\
3D segmentation & Square Attack & Black-box & Range-view tensor \\
3D detection & Point perturbation, removal, injection & White-box & Point cloud \\
3D detection & LiDAttack & Black-box & Point cloud \\
\bottomrule
\end{tabular}%
}
\end{table}

\section{Experimental Evaluation}

\subsection{Experimental Setup}
\label{subsec:experimental_setup}

\subsubsection{3D Object Detection}
\label{subsubsec:det_setup}
\textbf{Datasets.} We evaluate \shield{} on KITTI and a CARLA-based simulated dataset.
KITTI provides real LiDAR point clouds from a 64-channel sensor with
3D object annotations from urban driving scenes; all results are reported at the moderate
difficulty level. The CARLA dataset was
collected in Town06 over 3,000 frames at 10\,Hz using a 64-channel
LiDAR sensor with multiple vehicles in the scene, providing raw point
clouds, ground-truth poses, timestamps, and transformation matrices. \textbf{Evaluated Detectors.} We evaluate SECOND~\cite{Yan2018SENSORS}, a sparse convolutional
voxel-based detector, and PointPillars~\cite{Lang2019CVPR}, a pillar-based detector that encodes the point cloud into a pseudo-image representation.

\subsubsection{3D Semantic Segmentation}
\textbf{Datasets.} We evaluate on SemanticKITTI~\cite{Behley2019ICCV}, a large-scale urban driving dataset collected with a Velodyne HDL-64E sensor (range-view
size $64{\times}1024$, sequences 00--04 train, 05 validation, 06--08 test),
and SemanticPOSS~\cite{Pan2020SemanticPOSS}, a campus-scale dataset
acquired with a PANDORA 40-beam sensor (range-view size
$40{\times}1800$, sequences 0,1,4,5 train, 3 validation, 2 test). \textbf{Evaluated Segmentation Models.} We evaluate three pretrained range-view models: LENet~\cite{Feng2022LENet}, a lightweight encoder--decoder with multi-scale convolutional attention; CENet~\cite{Cheng2022TIP}, a projection-based model with large-kernel convolutions and auxiliary heads; and FIDNet~\cite{Zhao2021IROS}, a range-image model with a fully interpolation decoding module.

\subsection{Attack Configurations}
\label{subsec:attack_configurations}

\subsubsection{3D Object Detection}
\label{subsubsec:det_attack_config}
\textbf{White-box attacks.} For point perturbation, we apply an untargeted $\ell_2$-bounded PGD attack with $\epsilon \in \{5,7,10\}$\,cm, 40 iterations, step size
$\alpha=\epsilon/30$, and random initialization.
For point removal, we use saliency-driven greedy removal over 10
iterations at removal ratios $\{0.5,1.0,1.5\}\%$.
For point injection, we inject 300 points initialized in salient
regions and refined under $\ell_2$ constraints, with displacement
budget $\epsilon \in \{0.3,0.5,0.7\}$\,m. \textbf{Black-box attack.} We employ a LiDAttack-based~\cite{Chen2025LiDAttack} point
injection attack using only forward passes: a genetic
algorithm optimizes $K=20$ adversarial points within a
$0.5$\,m margin of the target bounding box, using at most
1000 forward queries per target, evaluated on SECOND on KITTI.

\subsubsection{3D Semantic Segmentation}
\label{subsubsec:sem_attack_config}\textbf{White-box attacks.} 
We apply PGD under both $\ell_2$ and $\ell_\infty$ constraints with $\epsilon \in \{3, 6, 9\}$ and  $\epsilon \in \{0.1, 0.3, 0.5\}$ respectively, $T=40$ iterations, and the step size $\alpha = \epsilon / T$. \textbf{Black-box attack.} We employ Square Attack~\cite{Andriushchenko2020SquareAttack} under an $\ell_\infty$ constraint with $\epsilon = 0.3$, 1000 queries, one random restart, and an initial perturbation probability $p_{\mathrm{init}} = 0.4$, evaluated on LENet on SemanticKITTI.

\subsection{Evaluation Metrics}
\label{subsec:evaluation_metrics}
\textbf{Detection Metrics}
\label{subsubsec:det_metrics}
Attack effectiveness is measured using the absolute mAP and the mAP ratio $(\mathrm{mAP}_{\mathrm{adv}} /
\mathrm{mAP}_{\mathrm{clean}}) \times 100$, where lower
values indicate stronger degradation. Geometric perceptibility is
measured by the symmetric Chamfer Distance (CD) between $P$
and $P_{\mathrm{adv}}$, with lower CD indicating better
geometric plausibility. \textbf{Segmentation Metrics}
\label{subsubsec:seg_metrics}
We report mean IoU (mIoU) averaged over all classes.

\subsection{Results and Discussion}
\label{subsec:results_discussion}
\subsubsection{3D Object Detection Results}
\label{subsubsec:det_results}
Table~\ref{tab:detection_ratio_cd}
reports mAP ratio and Chamfer Distance.
Both detectors degrade under adversarial manipulation on KITTI and
CARLA.
\textbf{Point perturbation} produces the strongest degradation. The mAP
ratio decreases consistently with the displacement budget on both
datasets, and the effect is particularly pronounced for PointPillars,
which drops to 10.59\% on KITTI and 3.46\% on CARLA at
$\epsilon=10$\,cm. SECOND retains higher relative performance,
suggesting that voxel-based sparse convolutional detectors are less
sensitive to coordinate-level perturbations than pillar-based encodings. \textbf{Point removal} leads to more gradual degradation. SECOND retains 67.98\% on KITTI and 85.32\% on CARLA, while
PointPillars retains 57.40\% and 80.02\% at the highest
removal ratio; the increasing Chamfer Distance reflects the
larger structural change caused by removing more points.
\textbf{Point injection} produces moderate degradation with lower Chamfer Distance values, as only a limited number of
synthetic points are added to the original cloud. 
On KITTI, increasing the attachment radius does not always
strengthen the attack: SECOND mAP ratio drops to 65.28\%
at $\epsilon=0.5$\,m but recovers to 71.12\% at
$\epsilon=0.7$\,m, with similar behavior for
PointPillars, suggesting that large budgets displace
injected points from the most influential regions. The black-box LiDAttack reduces SECOND mAP from 70.43 to 52.51 on KITTI (Table~\ref{tab:blackbox_unified}), corresponding to a relative drop of approximately 25\%.
\begin{table*}
\centering
\caption{Adversarial impact on 3D detection across KITTI and
CARLA: mAP ratio (\%) and Chamfer Distance (CD).}
\label{tab:detection_ratio_cd}
\scriptsize
\renewcommand{\arraystretch}{1.08}
\setlength{\tabcolsep}{3pt}
\resizebox{0.95\linewidth}{!}{%
\begin{tabular}{ll *{3}{cc} *{3}{cc} *{3}{cc}}
\toprule
& &
\multicolumn{6}{c}{\textbf{Point Perturbation:} $\epsilon$ (cm)} &
\multicolumn{6}{c}{\textbf{Point Removal: Drop Ratio (\%)}} &
\multicolumn{6}{c}{\textbf{Point Injection:} $\epsilon$ (m)} \\
\cmidrule(lr){3-8}\cmidrule(lr){9-14}\cmidrule(lr){15-20}
& &
\multicolumn{2}{c}{5} & \multicolumn{2}{c}{7} & \multicolumn{2}{c}{10} &
\multicolumn{2}{c}{0.5} & \multicolumn{2}{c}{1.0} & \multicolumn{2}{c}{1.5} &
\multicolumn{2}{c}{0.3} & \multicolumn{2}{c}{0.5} & \multicolumn{2}{c}{0.7} \\
\cmidrule(lr){3-4}\cmidrule(lr){5-6}\cmidrule(lr){7-8}
\cmidrule(lr){9-10}\cmidrule(lr){11-12}\cmidrule(lr){13-14}
\cmidrule(lr){15-16}\cmidrule(lr){17-18}\cmidrule(lr){19-20}
\textbf{Dataset} & \textbf{Detector} &
mAP Ratio & CD & mAP Ratio & CD & mAP Ratio & CD &
mAP Ratio & CD & mAP Ratio & CD & mAP Ratio & CD &
mAP Ratio & CD & mAP Ratio & CD & mAP Ratio & CD \\
\midrule
\multirow{2}{*}{KITTI}
& SECOND
& 90.91 & 0.00403 & 83.10 & 0.00703 & 65.48 & 0.01247
& 83.15 & 0.00255 & 74.18 & 0.00649 & 67.98 & 0.01533
& 71.73 & 0.00083 & 65.28 & 0.00195 & 71.12 & 0.00703 \\
& PointPillars
& 33.02 & 0.00399 & 19.56 & 0.00703 & 10.59 & 0.01252
& 80.25 & 0.01247 & 62.07 & 0.03597 & 57.40 & 0.06159
& 87.85 & 0.00014 & 75.83 & 0.00032 & 76.76 & 0.00057 \\
\midrule
\multirow{2}{*}{CARLA}
& SECOND
& 89.06 & 0.00466 & 82.80 & 0.00842 & 73.21 & 0.01533
& 95.28 & 0.02177 & 91.77 & 0.07035 & 85.32 & 0.19999
& 87.72 & 0.00054 & 78.46 & 0.00126 & 77.24 & 0.00226 \\
& PointPillars
& 22.82 & 0.00452 & 13.87 & 0.00823 & 3.46  & 0.01500
& 89.97 & 0.01924 & 85.23 & 0.05876 & 80.02 & 0.11100
& 95.93 & 0.00011 & 93.90 & 0.00021 & 89.72 & 0.00037 \\
\bottomrule
\end{tabular}
}
\end{table*}

\subsubsection{3D Semantic Segmentation Results}
\label{subsubsec:seg_results}

Table~\ref{tab:seg_l2_linf} reports LENet robustness under $\ell_2$ and
$\ell_\infty$ attacks on SemanticKITTI and SemanticPOSS. Across both datasets,
adversarial perturbations consistently reduce mIoU, with stronger degradation
as the perturbation budget increases. On SemanticKITTI, mIoU drops from 64.7
to 41.7 under the strongest $\ell_2$ attack and to 34.7 under the strongest
$\ell_\infty$ attack. Similar behavior is observed on SemanticPOSS, where mIoU
decreases from 52.7 to 34.4 under $\ell_2$ and to 28.8 under
$\ell_\infty$. 
The black-box Square Attack further reduces SemanticKITTI mIoU from 64.7 to 19.5 (Table~\ref{tab:blackbox_unified}), confirming vulnerability even without gradient access. The backbone comparison in Table~\ref{tab:dual_backbone_robustness} further indicates that LENet, CENet, and FIDNet all degrade under attack.

\begin{table}
\caption{Segmentation robustness on SemanticKITTI and SemanticPOSS.}
\label{tab:seg_l2_linf}
\centering
\scriptsize
\setlength{\tabcolsep}{4pt}
\renewcommand{\arraystretch}{1.05}
\resizebox{0.68\linewidth}{!}{
\begin{tabular}{l l c c c}
\toprule
\textbf{Dataset} & \textbf{Method}
& \multicolumn{3}{c}{\textbf{mIoU}} \\
\cmidrule(lr){3-5}

\multicolumn{5}{c}{$\ell_2$ attack} \\
\cmidrule(lr){3-5}
& & $\boldsymbol{\epsilon{=}3}$ 
  & $\boldsymbol{\epsilon{=}6}$ 
  & $\boldsymbol{\epsilon{=}9}$ \\
\midrule
\multirow{2}{*}{SemanticKITTI}
& No Attack & 64.7 & 64.7 & 64.7 \\
& Attack    & 50.1 & 46.9 & 41.7 \\
\midrule
\multirow{2}{*}{SemanticPOSS}
& No Attack & 52.7 & 52.7 & 52.7 \\
& Attack    & 39.8 & 36.5 & 34.4 \\

\midrule
\multicolumn{5}{c}{$\ell_\infty$ attack} \\
\cmidrule(lr){3-5}
& & $\boldsymbol{\epsilon{=}0.1}$ 
  & $\boldsymbol{\epsilon{=}0.3}$ 
  & $\boldsymbol{\epsilon{=}0.5}$ \\
\midrule
\multirow{2}{*}{SemanticKITTI}
& No Attack & 64.7 & 64.7 & 64.7 \\
& Attack    & 43.9 & 38.3 & 34.7 \\
\midrule
\multirow{2}{*}{SemanticPOSS}
& No Attack & 52.7 & 52.7 & 52.7 \\
& Attack    & 35.5 & 29.1 & 28.8 \\

\bottomrule
\end{tabular}
}
\end{table}

\begin{table}
\centering
\caption{Black-box robustness.}
\label{tab:blackbox_unified}
\scriptsize
\setlength{\tabcolsep}{4pt}
\renewcommand{\arraystretch}{0.95}
\begin{tabular}{lcccc}
\toprule
& \multicolumn{2}{c}{\textbf{Segmentation (SemanticKITTI)}}
& \multicolumn{2}{c}{\textbf{Detection (KITTI)}} \\
\cmidrule(lr){2-3}\cmidrule(lr){4-5}
& No Attack & Square Attack & No Attack & LiDAttack \\
\midrule
mIoU / mAP & 64.70 & 19.50 & 70.43 & 52.51 \\
\bottomrule
\end{tabular}
\end{table}


\begin{table}
\centering
\caption{Segmentation backbone robustness, SemanticKITTI ($\ell_2$, $\epsilon=6$).}
\label{tab:dual_backbone_robustness}
\scriptsize
\setlength{\tabcolsep}{5pt}
\renewcommand{\arraystretch}{1.05}
\begin{tabular}{lcccccc}
\toprule
& \multicolumn{2}{c}{\textbf{LENet}~\cite{Feng2022LENet}}
& \multicolumn{2}{c}{\textbf{CENet}~\cite{Cheng2022TIP}}
& \multicolumn{2}{c}{\textbf{FIDNet}~\cite{Zhao2021IROS}} \\
\cmidrule(lr){2-3}\cmidrule(lr){4-5}\cmidrule(lr){6-7}
& Clean & Attack & Clean & Attack & Clean & Attack \\
\midrule
SemanticKITTI, $\epsilon=6$
& 64.70 & 46.90
& 61.70 & 44.30
& 60.80 & 44.10 \\
\bottomrule
\end{tabular}
\end{table}

\section{Conclusion}
\label{sec:conclusion}


We presented \shield{}, a unified and fully Dockerized
framework for adversarial stress-testing of LiDAR-based 3D
perception models. Unlike library-centric toolkits, \shield{}
decouples the assessment engine from the target model through
a minimal REST interface, requiring no manual environment
reconstruction or model-specific glue code. The framework
supports both 3D semantic segmentation and 3D object detection
through the same workflow, exposing a task-aware suite of
state-of-the-art attacks: PGD and Square Attack for
segmentation, and point perturbation, removal, injection, and
LiDAttack for detection. Experimental results on
SemanticKITTI, SemanticPOSS, KITTI, and CARLA confirm that
the integrated attacks expose significant vulnerabilities
across heterogeneous LiDAR backbones. 

\balance
\bibliographystyle{IEEEtran}

\end{document}